# Attention-based Neural Network for Driving Environment Complexity Perception

Ce Zhang, *Student Member, IEEE*, Azim Eskandarian, *Senior Member, IEEE,* Xuelai Du

*Abstract*—Environment perception is crucial for autonomous vehicle (AV) safety. Most existing AV perception algorithms have not studied the surrounding environment complexity and failed to include the environment complexity parameter. This paper proposes a novel attention-based neural network model to predict the complexity level of the surrounding driving environment. The proposed model takes naturalistic driving videos and corresponding vehicle dynamics parameters as input. It consists of a Yolo-v3 object detection algorithm, a heat map generation algorithm, CNN-based feature extractors, and attention-based feature extractors for both video and time-series vehicle dynamics data inputs to extract features. The output from the proposed algorithm is a surrounding environment complexity parameter. The Berkeley DeepDrive dataset (BDD Dataset) and subjectively labeled surrounding environment complexity levels are used for model training and validation to evaluate the algorithm. The proposed attention-based network achieves 91.22% average classification accuracy to classify the surrounding environment complexity. It proves that the environment complexity level can be accurately predicted and applied for future AVs' environment perception studies.

## I. INTRODUCTION

Autonomous Vehicles (AVs) are rapidly developing and play an important role in the future of intelligent transportation systems [1]. An AV's main tasks are environment perception, path planning, and vehicle control to achieve autonomous driving [2]. Among the three tasks, perception is the first but vital step, directly affecting the path planning and vehicle control performance. Therefore, a comprehensive and precise surrounding environment perception algorithm is crucial for AVs' safety.

Currently, AV perception methodologies can be categorized into (a) mediated perception, (b) behavior reflex perception, and (c) direct perception, as shown in Fig 1 [3]. Mediated perception (MP) is the most popular approach that involves applying an object detection algorithm to recognize driving-relevant objects. After object recognition, the output features are employed for further path planning and vehicle control. The behavior reflex perception (BRP) methodology directly maps sensory inputs to driving action through deep learning methods. Direct perception (DP), proposed by C. Chen et al. falls in between (a) and (b) [4]. Instead of detecting objects or directly predicting driving actions, the DP method predicts the affordance for driving actions (vehicle angle, position, distance, etc.) from the sensor inputs. Then, driving affordance results are used for driving action generation.

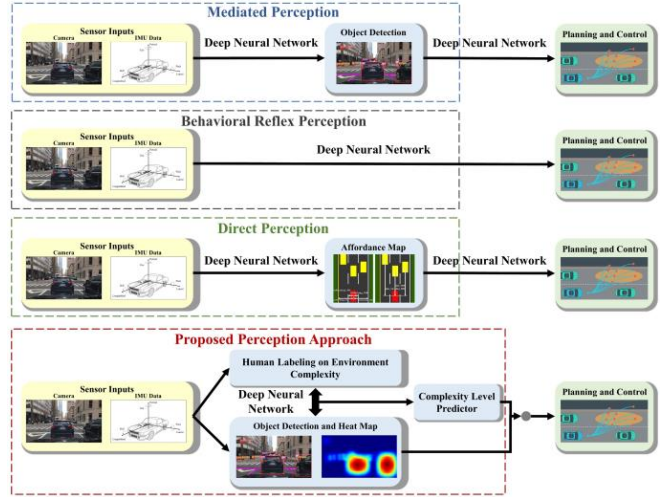

Figure 1. AVs Popular Perception Approaches

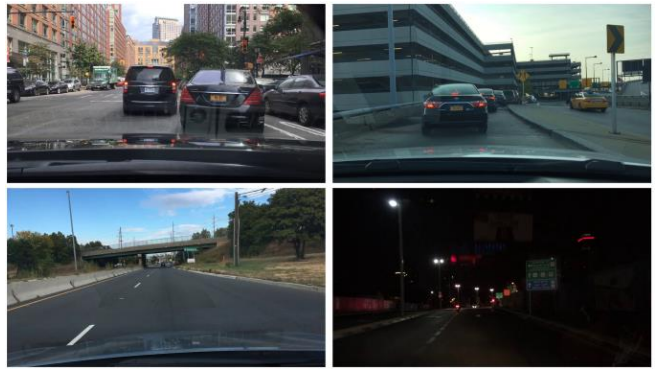

Figure 2. Selected Naturalistic Driving Videos from the BDD Dataset. By comparing the top left and right figures, even though the weather condition, speed condition, and the number of objects are similar, most drivers agree that the top left driving scenario is more complex because the vehicles are covering the front road view and some drivers suspect that there might be pedestrians crossing the road. By comparing the bottom left and right figure, both scenarios contain small objects. However, the bottom right figure surrounding environment is more complex than the bottom left because the bottom right visibility is poor. Existing perception algorithms do not take these human insights into account.

Ce Zhang is with Virginia Tech Department of Mechanical Engineering, Blacksburg, VA, 24061, USA (phone: 540-998-1186; e-mail: zce@vt.edu). He is a Ph.D. student at Virginia Tech Autonomous Systems and Intelligent Machines (ASIM) Lab.

Azim Eskandarian is the Department Head and Nicholas and Rebecca Des Champs Chair/Professor of the Mechanical Engineering Department and the Director of ASIM Lab, Virginia Tech, Blacksburg, VA, 24061, USA

Xuelai Du is with Virginia Tech Department of Mechanical Engineering, Blacksburg, VA, 24061, USA. He is a undergraduate researcher at Virginia Tech ASIM Lab.

All existing methods can provide accurate perception information for further path planning and vehicle control. However, these approaches do not take surrounding environment complexity into account. Surrounding environment complexity cannot be simply described by just the number of objects or vehicle performances. It is a comprehensive perception process that mimics humans' understandings and forecasts of the surrounding environment. Therefore, adding the surrounding environment complexity parameter to the perception model will enable the algorithm to sense the surrounding environment like a human being (Figure 2), which lets the autonomous system operate the vehicle more like a human driver. An accurate surrounding environment complexity parameter prediction can enable the autonomous vehicle to drive safer. Also, a similar-to-human autonomous vehicle driving style is more friendly to most users.

In this paper we asked volunteers to label and categorize surrounding environment complexity into five classes by watching naturalistic driving videos. Then, we developed a novel attention-based neural network model to predict the volunteer-labeled surrounding environment complexity results. Based on the validation results, classification accuracy of the proposed algorithm is 91.22%. This result proves that the proposed algorithm can precisely predict the surrounding environment complexity level.

The remaining of the paper is organized as follows. Section II reviews recent works on object detection algorithms for AV perception. Section III discusses the proposed attention-based neural network, while section IV presents the open-source naturalistic driving dataset used for this study. Section V presents the results, and finally, section VI concludes our current work and discusses potential improvements to this study and future research.

## II. RELATED WORKS

This section only briefly discusses our literature review on popular object detection algorithms for AV perception due to space limitation. Most object detection algorithms learn image features through a deep learning (neural network) method. Since object detection requires object localization and object type classification, the learned features can be categorized into one-stage and two-stage object detection algorithms [5].

### A. One-stage Object Detection Algorithms

One-stage object detection algorithms generate object location and object classification results directly in one stage. These detection algorithms do not require a region proposal process, which is usually simpler and faster than two-stage detection algorithms. In what follows, single-shot multi-box detector (SSD) [6] and you only look at once algorithm (Yolo) [7] are explained.

*1) SSD Detection Algorithm*

The SSD algorithm was proposed by W. Liu, et al. in 2015. It uses a VGG16 network [8] as the backbone for spatial feature extraction. The SSD algorithm is composed of 6 stages of hierarchical feature extractions, and has two input versions: SSD-300 (input image size is transformed to $[300 \times 300]$) and SSD-512 (input image size is transformed to $[512 \times 512]$). Both versions were evaluated on both the VOC 2007 and the COCO dataset. For the VOC 2007 dataset, the SSD-300 mean average precision (mAP) was 74.3% with a frame rate of 46 fps, and the SSD-512 mAP was 76.8 with a frame rate of 19 fps. For the COCO dataset, the mAP was 43.1 and 48.5 for SSD-300 and 512, respectively. In 2017, the SSD algorithm was further improved by employing several deconvolutional layers after hierarchical feature extractions, namely Deconvolutional SSD (DSSD), detailed in [9].

*2) Yolo Detection Algorithm*

Yolo detection algorithm is a series of real-time object detection algorithms that upgraded from v1 to v5. For autonomous vehicle applications, the most popular algorithm version is Yolo-v3 because of the model's compact size and ease of application [10]. The backbone of the Yolo detection algorithm is the Darknet. Compared to Yolo-v1, Yolo-v3's backbone is upgraded to darknet-53, a 53 layers convolutional neural network model. Furthermore, the object localization classifier is logistic regression, and the object class prediction is upgraded to independent logistic classifiers. The Yolo-v3 mAP on the COCO dataset is similar to the SSD network, while the FPS is three times faster than the SSD.

### B. Two-stage Object Detection Algorithms

Two-stage object detection algorithms conduct a region proposal process and then classify the object in the proposed region. Next, the faster-RCNN [11] and mask-RCNN [12] are explained.

*3) Faster R-CNN*

Faster R-CNN is proposed by S. Ren et al. in 2016 to improve the processing speed of fast R-CNN [13] and the R-CNN algorithm. The backbone of faster R-CNN is deep CNN (DCNN). To improve computational speed, faster R-CNN employs a region proposal algorithm that reduces the number of object region proposals and improves the region proposal quality. Faster R-CNN was evaluated on the VOC 2007 dataset, and the mAP was 59.9 with a frame rate of 17 fps.

*4) Mask R-CNN*

Mask R-CNN further enhances faster R-CNN by improving the overall detection accuracy and small target detection. The main improvement in mask R-CNN is the pooling layer. For mask R-CNN, ROIAlign pooling layer was employed instead of the regular ROI pooling layer. Mask R-CNN was evaluated on the COCO dataset with a mAP of over 60.

## III. METHODS

The overall workflow for the proposed algorithm (shown in Figure 1) input is comprised of two categories: (a) camera input and (b) vehicle dynamic parameters input. Camera input is an explicit measurement for the surrounding environment complexity because it records obstacles and objects during vehicle driving. Vehicle dynamic parameters can be considered as implicit surrounding environment complexity measurements because they describe vehicle performance under specific driving scenarios. In this study, we use four essential vehicle dynamic parameters: longitudinal acceleration ($a_x$), speed ($v$), lateral acceleration ($a_y$), and yaw rate ($\dot{\varphi}_y$). $a_x$ and $v$ represent vehicle performance in the longitudinal direction while the latter two parameters describe it in the lateral direction. These inputs are passed into a series of feature extractors algorithms. Then, the outputs from each extractor are fused together for driving environment complexity classification. In the following sections, we discuss the details of the camera feature extractor and vehicle dynamic feature extractor, respectively.

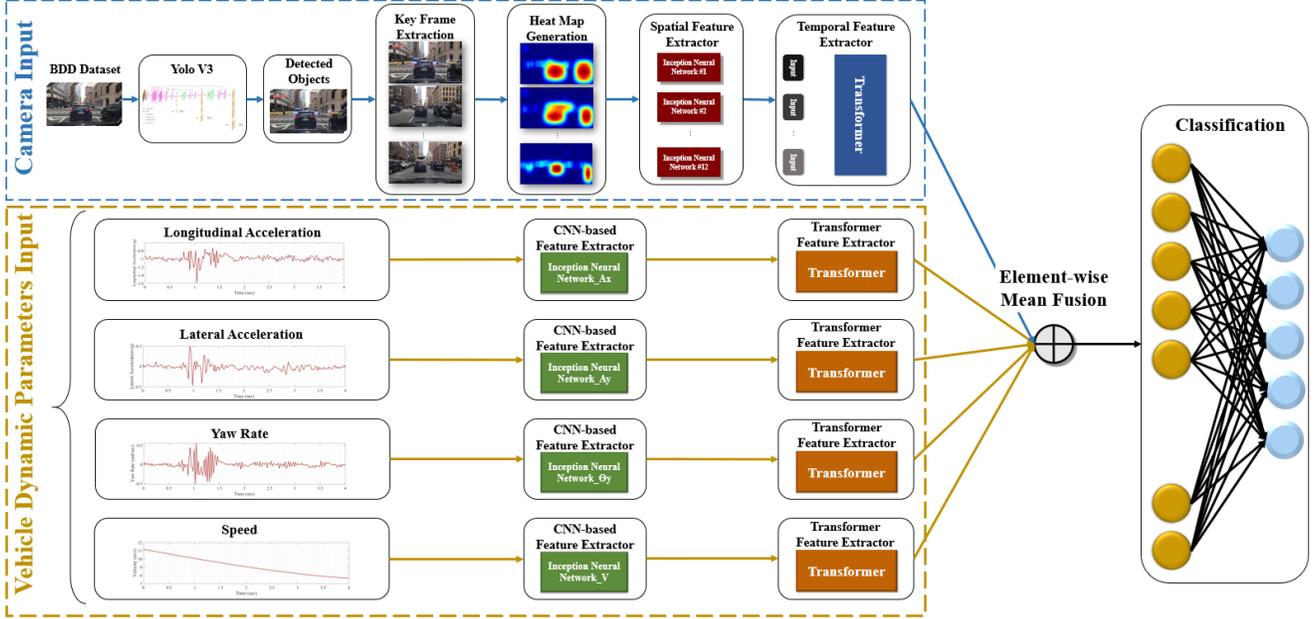

Figure 3. Overall workflow of the proposed algorithm. Four vehicle dynamic parameters and naturalistic driving videos are used as inputs.

## A. Camera Feature Extractor

The camera feature extractor algorithm employs the naturalistic driving videos as input and outputs a latent feature vector for data fusion and classification. The videos are fed into the Yolo-v3 algorithm for object detection at first. Then, twelve detected object keyframes are extracted for heat map generation. Each heat map image is used as the input for a spatial feature extractor and temporal feature extractor. Details about the Yolo-v3 object detection algorithm are explained in [10]. In this part, we explain the heatmap generation equation, spatial feature extractor, and temporal feature extractor in detail in the remainder of this paragraph.

### 1) Heat Map Generation

The heat map generation step's objective is to enhance the detected object features and reduce the proposed algorithm computational load. Class activation map (CAM) [14] is an effective solution for detected object heat map generation by extracting the feature map of the last layer from the feature extractor (usually CNN). However, the CAM method can only generate a heat map with the same intensity level for all detected objects, which cannot distinguish object importance according to different object classes. Therefore, we developed a novel heat map generation method based on the bounding box output from an object detection algorithm, as shown in TABLE I and Fig 2. When employing the novel heat map generation algorithm, the heat intensity value is varied based on both detected object sizes and object classes.

### 2) Spatial Feature Extractor

The backbone of the spatial feature extractor is a CNN-based neural network, as shown in Figure 3. Since Yolo-v3 has already detected the objects in the keyframe images, this spatial feature extractor's objective is to extract more explicit features such as "heat regions" sizes, positions, and intensity values. According to [15], the inception module is an effective and fast spatial feature extraction algorithm because it extracts the spatial features with different kernel sizes in parallel. Therefore, we employ a similar inception module as the spatial feature extractor. A residual module is employed to prevent overfitting and reduce the learning degradation problem [16]. In the proposed algorithm, we select four convolutional layers with kernel sizes of $[3 \times 1], [1 \times 3], [3 \times 3]$, and $[5 \times 5]$, and the spatial feature extractor output size is $[1 \times 32 \times 18]$.

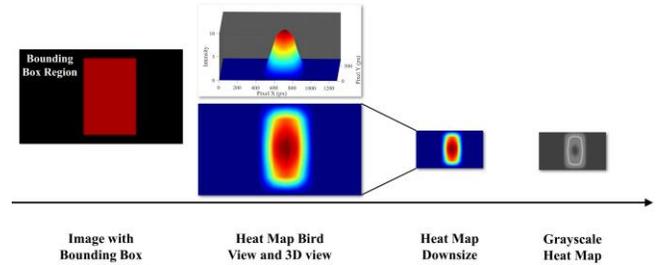

Figure 4. Heat Map Generation Procedures

### 3) Temporal Feature Extractor

The temporal feature extractor is a self-attention-based neural network (transformer), as shown in Figure 4. Vaswani initially proposed the transformer model for natural language processing studies [17]. Dosovitskiy has extended the transformer model for image recognition studies [18]. The transformer model's benefits are (a) fast processing speed and (b) process all input at one time. Since the proposed algorithm employs video extracted keyframes as input, processing all keyframes at one time allows the model to learn the temporal features' variation among different keyframes due to vehicle motion.

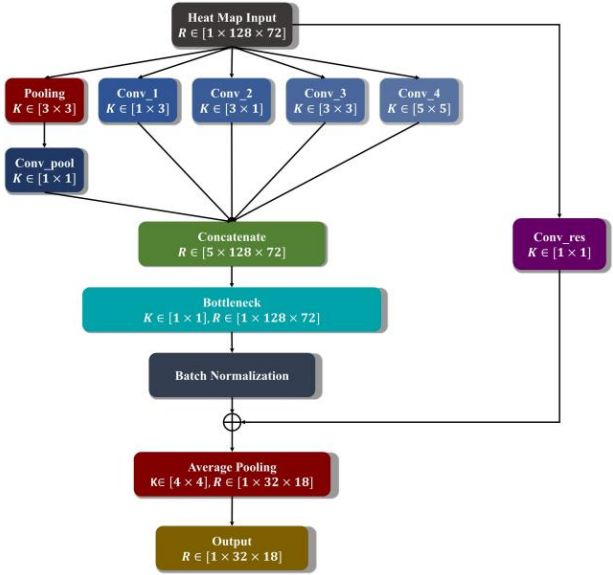

Figure 5. CNN-based Spatial Feature Extractor. $R$ represents the dimension of the output data from each layer, and $K$ represents the kernel size

TABLE I.    PSEUDOCODE FOR HEAT MAP GENERATION

```
Algorithm 1: Heat Map Generation
1:  Bounding Boxes Localization
2:    (X, classes) = Yolo_v3(image), [x_lb, x_rt, y_lb, y_rt] = X
3:  For i in range(size(X))
4:    Generate a Rectangle with Bounding Box Shape
5:    [X_i, Y_i] = grid(x_lb[i]: x_rt[i], y_lb[i]: y_rt[i])
6:    Linear Interpolation for Horizontal Pixels X_i
7:    a_0 = -π/2, a_1 = π/2, b_0 = min(X_i), b_1 = max(X_i)
8:    x_h = (x_in - b_0)·(a_1 - a_0)/(b_1 - b_0) + a_0 → (x_dim ∈ [W × 1])
9:    Extend the Horizontal Pixel Dimension
10:   x_h = extend(x_h) → (x_dim ∈ [W × H])
11:   Linear Interpolation for Vertical Pixels Y_i
12:   a_0 = -π/2, a_1 = π/2, b_0 = min(Y_i), b_1 = max(Y_i)
13:   y_v = (y_in - b_0)·(a_1 - a_0)/(b_1 - b_0) + a_0 → (x_dim ∈ [1 × H])
14:   Extend the Vertical Pixel Dimension
15:   y_b = extend(y_v) → (y_dim ∈ [W × H])
16:   Weight Factor Determination based on Yolo-v3 Detected Classes
17:   For j in range(size(classes))
18:     if classes[i] is pedestrian and cyclist
19:       n = 4
20:     else if classes[i] is vehicle, traffic sign, and traffic light
21:       n = 2
22:     else
23:       n = 1
24:   Heat Intensity Value Generation for Horizontal (x_h) and Vertical (y_v)
25:   z_h[i][j] = exp(√n(j) · cos(x_h)), z_v[i][j] = exp(√n(j) · cos(y_v))
26:   L2 Distance Summation for Overall Heat Intensity Value
27:   z_total[i][j] = √(z_h²[i][j] + z_v²[i][j])
28: Sum All Intensity Value for the Image
29: Z = Σ z_total[i][j]
```

The proposed transformer model has twelve inputs, which are the outputs from the previous inception modules. The inputs are sorted based on timestep and processed altogether to the transformer model. The transformer model contains three parts: normalization module, self-attention module, and feedforward module.

The normalization module simply adds the input data to the output data and normalizes the sum of the input and output data to prevent overfitting and gradient degradation. As for the self-attention module, the output is a set of self-attention scores, which is achieved by calculating

$$Z_{score} = \left(\frac{Q_{out} \times K_{out}^T}{\sqrt{d}}\right) \times V_{out} \quad (1)$$

where $Q_{out}$, $K_{out}$, and $V_{out}$ are the outputs from three transformation matrices, namely query (Q), key (K), and value (V). The weights of Q, K, and V are training parameters in the transformer neural network. $\sqrt{d}$ is a factor of dimension reduction, which is set to 8 in this algorithm. The output matrices $Q_{out}$, $K_{out}$, and $V_{out}$ are obtained through

$$Q_{out} = X_{in} \times Q \quad (2)$$
$$K_{out} = X_{in} \times K \quad (3)$$
$$V_{out} = X_{in} \times V \quad (4)$$

where $X_{in}$ is the input data, and $Q$, $K$, and $V$ matrices are the transformation matrices.

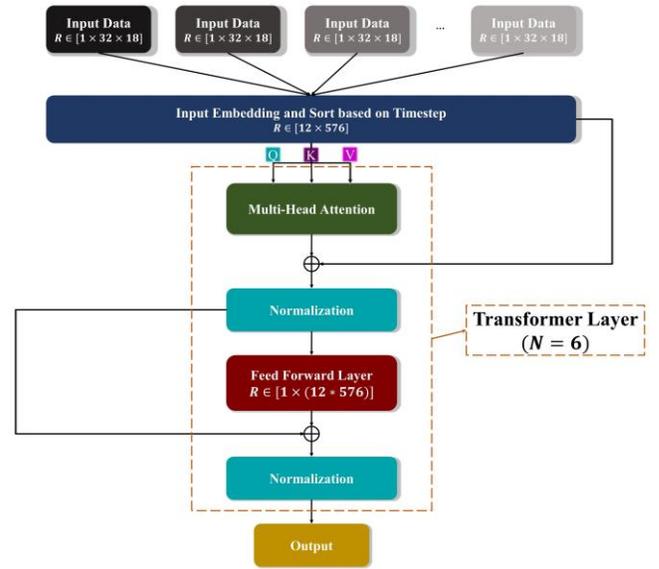

Figure 6. Transformer Model Architecture for Camera Feature Extractor

The feedforward module is a multi-layer perceptron (MLP) network to transform the self-attention scores for either further encoding or classification. In the proposed transformer model, the number of MLP layers is set to be three.

According to an ablation study, the number of layers for the transformer model is six. For the first five transformer modules, the feedforward module output dimension is the same as the input, while the sixth transformer module's output dimension is set to be the same as the dynamic parameters' transformer network output ($R \in [1 \times 200]$) for data fusion.

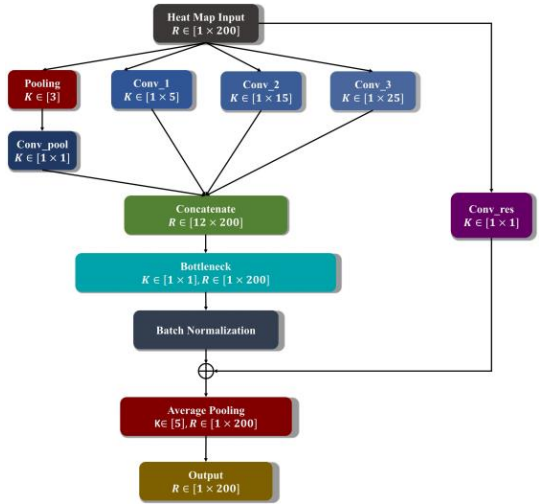

Figure 7. CNN-based Feature Extractor for Vehicle Dynamic Parameters

### B. Vehicle Dynamic Feature Extractors

Vehicle dynamic feature extractor architectures are similar to the camera feature extractor, and are comprised of a CNN-based feature extractor and a self-attention-based neural network model. The objective of the CNN-based feature extractor is to remove noises and artifacts from the time-series signals. C. Zhang et al. have proved that CNN can filter noise signals and extract features effectively for time-series signals processing [19]. Since the input data is 1D time-series signals, the vehicle dynamic feature extractor kernel matrices (CNN kernel matrices and transformer transformation matrices) are all one-dimensional matrices. The CNN-based feature extractor and the transformer model architecture are shown in fig. 6 and 7, respectively.

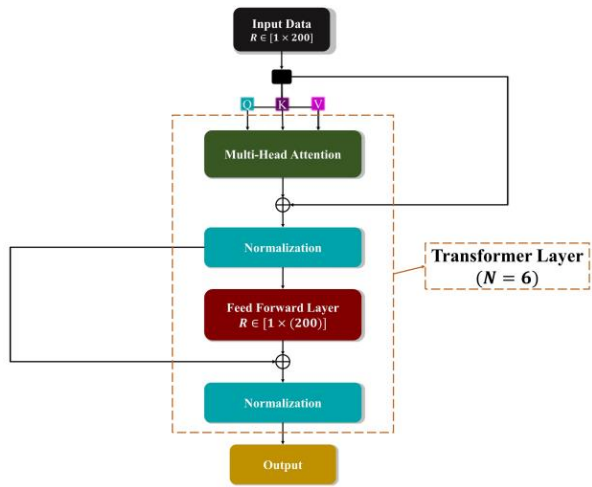

Figure 8. Attention-based Neural Network Architecture for Vehicle Dynamic Parameters

## IV. DATASET AND SUBJECTIVE LABELING

To verify and evaluate the proposed model performance, the Berkeley DeepDrive (BDD) Dataset was used [20]. Furthermore, we hired volunteers to watch the BDD dataset driving videos and label the driving environment complexity based on their previous driving experiences.

### A. Berkeley DeepDrive Dataset

The BDD Dataset is one of the largest naturalistic driving video datasets for transportation safety and autonomous vehicle perception. The dataset contains 100,000 driving videos from New York, Bay Area, San Francisco, and Berkeley regions. Each video length is around 40 seconds, and the frame rate is 30 fps.

Since labeling driving environment complexity is time-consuming for volunteers, we only randomly selected 800 videos from the whole dataset. Since the 40-second video length is too long to describe the surrounding driving environment complexity, we further divide the 40-second video into ten samples, which is 4 seconds video length per sample. After removing failed video samples, the total number of driving videos was 786 and the total number of samples was 7,860, as shown in TABLE II.

TABLE II. TRAINING AND TESTING SAMPLES SUMMARY TABLE

| Classes | Total Sample | Percentage (%) | Train Sample | Validate Sample |
|---|---|---|---|---|
| 0 | 2581 | 32.84 | 2055 | 526 |
| 1 | 3523 | 44.83 | 2819 | 704 |
| 2 | 1461 | 18.59 | 1179 | 282 |
| 3 | 254 | 3.23 | 202 | 52 |
| 4 | 41 | 0.52 | 33 | 8 |
| Samples | 7860 | 100% | 6288 | 1572 |

Besides driving videos, the BDD dataset also provides IMU and GPS data accordingly. The IMU data contains important vehicle dynamic parameters: longitudinal, lateral, and vertical acceleration, and roll, yaw, pitch rate. The GPS data provides the vehicle speed, and the latitude and longitude coordinates of the vehicle. For our research, we only selected longitudinal acceleration, lateral acceleration, yaw rate, and vehicle speed as the dynamic parameters input because these are more related to vehicle performance and driving safety compared with the other parameters. Even though roll rate is important to vehicle safety [21], we ignored it as well for this study. The reason is that there is no commercial truck or passenger bus used as a testing vehicle. Therefore, dangerous roll behavior (rollover) does not tend to happen in the provided dataset.

### B. Driving Environment Complexity Labeling

Volunteers were hired for the driving environment complexity labeling work. The label score ranged from 0 to 4, where 0 represents the least complex (easy), and 4 represents the most complex (most difficult). An example of the volunteer's labeling rubric is shown in Table II. To ensure labeling consistency, each volunteer needed to watch all videos at least three times. The objective for the first time watching was to let volunteers have a general sense of the surrounding environment complexity for all naturalistic driving videos. The second time watching required volunteers to label the complexity level. The third time was to let volunteers validate their labeling results and ensure their scores were as consistent as possible.

TABLE III. VOLUNTEERS SURROUNDING ENVIRONMENT COMPLEXITY LABELING TABLE

| Video Name: 0000xxxx-xxxxxxxx | Complexity Level Score | Vehicle (Driving is 1 and Stop is 0) |
|---|---|---|
| 00:00-00:04 | 1 | 1 |
| 00:04-00:08 | 2 | 1 |
| 00:08-00:12 | 2 | 1 |
| 00:12-00:16 | 1 | 1 |
| 00:16-00:20 | 0 | 1 |
| 00:20-00:24 | 0 | 1 |
| 00:24-00:28 | 3 | 1 |
| 00:28-00:32 | 0 | 0 |
| 00:32-00:36 | 0 | 0 |
| 00:36-00:40 | 0 | 0 |

## V. RESULTS AND DISCUSSIONS

The proposed algorithm results can be separated into three categories: Yolo-v3 object detection performance, heat map generation results, and surrounding environment complexity classification accuracy.

### A. Yolo-v3 Object Detection Algorithm Performance

For the BDD dataset, the Yolo-v3 algorithm processing speed was 42.9 fps, and the mean average precision (mAP) was 80.52% [22]. We do not quantitatively discuss the Yolo-v3 classification accuracy because the Yolo-v3 algorithm is well-known among researchers. Its evaluation results can be searched for both in the original paper and several NN learning websites. A qualitative observation of the Yolo-v3 algorithm performance in different weather or time conditions is presented in this section. The potential influences of the Yolo-v3 algorithm performance on the surrounding environment complexity classification results are illustrated.

Fig. 9 presents the Yolo-v3 object detection results for different driving conditions. Yolo-v3 exhibits good performance in the daytime and clear weather conditions. Based on the bottom left detection results, the Yolo-v3 algorithm is still functional and presents accurate detections at strong sunlight reflection conditions. Moreover, the Yolo-v3 algorithm detection accuracy is also acceptable on rainy days during the daytime. Based on the right top and right bottom images, most vehicles and traffic signs can be detected. However, the performance is dramatically decreased at nighttime, especially for a rainy night condition. Based on the middle top and bottom figure, most on-road vehicles are not detected.

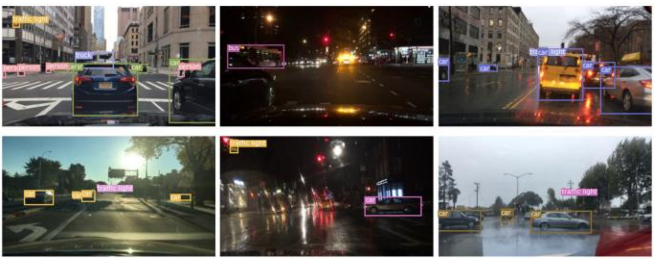

Figure 9. Yolo-v3 Detection Results on BDD Dataset

The object detection algorithm performance can directly affect the surrounding environment complexity classification. Since the inputs for the surrounding environment complexity feature extractor are heat maps, a poor object detection result can easily confuse the model because the heat maps' intensity is zero, as shown in Fig. 10. In Fig. 10, both heat map intensity values are close to 0, but the volunteers' labels are 0 and 3, respectively.

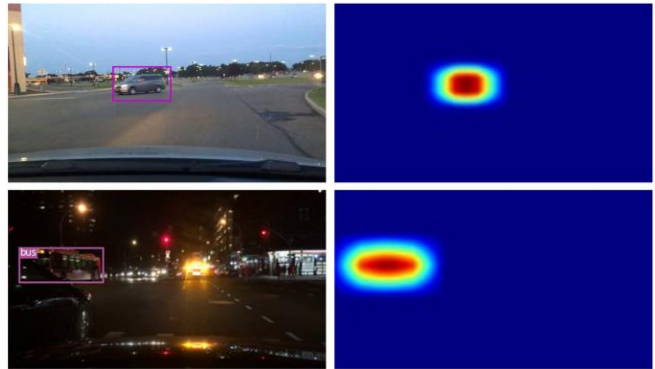

Yolo-v3 Detection Results at Different Driving Scenarios. The heat map intensity values are similar for both scenarios. However, for the complexity label, the left top figure is 0, while the left bottom figure is 3.

### B. Heat Map Generation Results

Fig. 11 presents an example of the heat map generation results. According to this figure, the proposed heat map generation algorithm successfully reflects the detected objects' size, position, and intensity value for different object classes. Based on the Yolo-v3 detection results, at least four vehicles and two persons are detected in Fig. 11. For heat map generation results, the heat intensity location and heat region size are the same as the Yolo-v3 detected bounding boxes results. Furthermore, the weight factor introduced in the heat generation algorithm successfully distinguishes different object classes. Even though the detected pedestrian bounding box size (Fig. 11) is smaller than the red SUV and the silver car, the heat intensity value is still higher than those vehicle classes because we set the weight on human classes higher than vehicle classes (See TABLE I about the weight factor details).

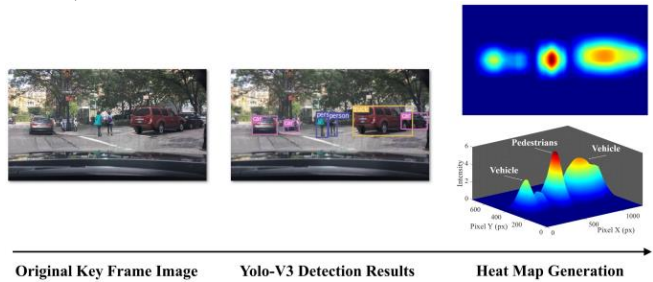

Figure 10. Heat Map Generation Results Example

### C. Surrounding Environment Complexity Classification Accuracy

According to the 5-fold cross-validation test, the average surrounding environment complexity classification accuracy is 91.22%, and classification accuracy results among different classes are shown in Table IV. As mentioned in the experiment dataset section, Class 0 indicates the lowest complexity while class 4 indicates the highest complexity level. According to the classification table, classification accuracy for classes 3 and 4 is much lower than classes 0 and 1. The reason is that the number of training and validation samples for "class 1" and "class 2" is much higher than the number of samples for "class 3" and "class 4", as shown in Table IV. Such bias in the sample size is reasonable for naturalistic driving studies where most of the driving scenarios are safe and relatively simple, while rare driving scenarios are dangerous and complex. To overcome this issue, we can further increase the sample size.

TABLE IV.  PROPOSED ALGORITHM CLASSIFICATION ACCURACIES AND CONFUSION MATRIX AMONG ALL CLASSES

|         | Class 0 | Class 1 | Class 2 | Class 3 | Class 4 |
|---------|---------|---------|---------|---------|---------|
| Class 0 | 482     | 24      | 5       | 0       | 1       |
| Class 1 | 30      | 659     | 28      | 9       | 1       |
| Class 2 | 11      | 21      | 245     | 1       | 0       |
| Class 3 | 3       | 0       | 4       | 42      | 0       |
| Class 4 | 0       | 0       | 0       | 0       | 6       |
| Acc (%) | 91.64   | 93.61   | 86.88   | 80.77   | 75.00   |

## VI. CONCLUSIONS AND FUTURE WORKS

This paper introduced a novel attention-based algorithm to process and classify the surrounding driving environment complexity. Based on the classification results, the proposed algorithm can successfully mimic human thoughts (classification) and classify the driving environment complexity into five classes. The average classification accuracy achieves 91.22% for the 5-fold cross-validation test.

However, there are still areas to improve the proposed algorithm. The first is heat map generation algorithm speed. Currently, the proposed heat map generation equation is too complex, which increases the computation load of the overall neural network. To ensure real-time processing, we need to simplify the equation to improve the processing speed. The second area is introducing object depth features. Currently, we employ vehicle dynamics parameters and surrounding object size and shape features. The surrounding object distance feature, which is another critical factor for driving safety and surrounding environment complexity, is not considered. The distance feature can be included through either stereovision camera or LiDAR sensors. The third area is extending the current algorithm to driving action. Predicting the surrounding environment complexity level is the first step. Developing an algorithm that links the environment complexity to a specific driving action is the key to improve AV driving safety.

In summary, we believe the study of surrounding environment complexity can enhance AVs' perception accuracy and precision. Incorporating surrounding environment complexity level into AV perception algorithms can allow future AVs to be more intelligent and safer.